\documentclass[11pt]{article}
\usepackage{times,latexsym}
\usepackage{url}
\usepackage[T1]{fontenc}
\usepackage{rotating}
\usepackage{array}
\usepackage{graphicx}
\usepackage[normalem]{ulem}

\usepackage[acceptedWithA]{style-files/TACL/tacl2021v1}

\usepackage{xspace,mfirstuc,tabulary}

\newif\iftaclinstructions
\taclinstructionsfalse 
\iftaclinstructions
\renewcommand{\confidential}{}
\renewcommand{\anonsubtext}{(No author info supplied here, for consistency with
TACL-submission anonymization requirements)}
\newcommand{\instr}
\fi

\iftaclpubformat 

\else

\fi

\usepackage{microtype}
\usepackage{graphicx}
\usepackage{subcaption}
\usepackage{booktabs} 
\usepackage{tikz} 
\usepackage{tikz-layers} 
\usepackage{xcolor,colortbl} 
\usepackage{amsmath} 
\usepackage{amssymb} 
\usepackage{wasysym} 
\usepackage{array}
\usepackage{times}
\usepackage{latexsym}

\usepackage{pgfplots}
\pgfplotsset{compat=newest} 

\usepackage{pifont} 
\newcommand{\cmark}{\ding{51}} %
\newcommand{\xmark}{\ding{55}} %
\usepackage{arydshln}
\usepackage{multirow}

\usepackage[T1]{fontenc}

\usepackage[utf8]{inputenc}

\usepackage{fancyhdr}

\newcommand{\tv}{\textsc{T5}}
\newcommand{\longtv}{\textsc{LongT5}}
\newcommand{\etoe}{\textsc{E2E}}
\newcommand{\aplan}{\textsc{Multitask}}
\newcommand{\iterative}{\textsc{Iterative}}
\newcommand{\twostage}{\textsc{2-Stage}}

\definecolor{Gray}{gray}{0.85}
\definecolor{LightCyan}{rgb}{0.0, 0.5, 1.0}
\definecolor{VeryLightCyan}{rgb}{0.0, 0.8, 1.0}
\definecolor{ForestGreen}{RGB}{34,139,34}
\definecolor{skyblue}{rgb}{0.53, 0.81, 0.92}

\definecolor{DarkBP}{RGB}{109, 157, 235}
\definecolor{MidBP}{RGB}{164, 194, 244}
\definecolor{LightBP}{RGB}{201, 218, 248}

\definecolor{forestgreen}{HTML}{009B55}
\definecolor{sepia}{HTML}{671800}
\definecolor{midnightblue}{HTML}{006795}
\definecolor{orangered}{HTML}{E24C00}
\definecolor{bblue}{HTML}{4F81BD}
\definecolor{rred}{HTML}{C0504D}
\definecolor{ggreen}{HTML}{9BBB59}
\definecolor{ppurple}{HTML}{9F4C7C}

\hypersetup{
           breaklinks=true,   
           colorlinks=true,   
           pdfusetitle=true,  
        }
\usepackage{breakurl}

\title{Conditional Generation with a Question-Answering Blueprint}

\author{
Shashi Narayan, Joshua Maynez, Reinald Kim Amplayo, Kuzman Ganchev, \\
\textbf{Annie Louis, Fantine Huot, Anders Sandholm, Dipanjan Das, Mirella Lapata} \\ Google Research \\ 
\texttt{\small shashinarayan@google.com, joshuahm@google.com, reinald@google.com,}\\ \texttt{\small kuzman@google.com,  annielouis@google.com, fantinehuot@google.com,} \\ 
\texttt{\small sandholm@google.com, dipanjand@google.com,
lapata@google.com}
}

\begin{document}
\maketitle

\begin{abstract}
  The ability to convey relevant and faithful information is critical
  for many tasks in conditional generation and yet remains elusive for
  neural seq-to-seq models whose outputs often reveal hallucinations
  and fail to correctly cover important details. In this work, we
  advocate planning as a useful intermediate representation for
  rendering conditional generation less opaque and more grounded.  We
  propose a new conceptualization of text plans as a sequence of
  question-answer (QA) pairs and enhance existing datasets (e.g., for
  summarization) with a QA \emph{blueprint} operating as a proxy for
  content selection (i.e.,~what to say) \emph{and} planning (i.e.,~in
  what order). We obtain blueprints automatically by exploiting
  state-of-the-art question generation technology and convert
  input-output pairs into input-blueprint-output tuples. We develop
  Transformer-based models, each varying in how they incorporate the
  blueprint in the generated output (e.g., as a global plan or
  iteratively). Evaluation across metrics and datasets demonstrates
  that blueprint models are more factual than alternatives which do
  not resort to planning and allow tighter control of the generation
  output.
\end{abstract}

\section{Introduction}
\label{sec:introduction}


Neural generation models are often prone to
hallucination
\cite{song-etal-2018-structure,maynez-etal-2020-faithfulness,kryscinski-etal-2020-evaluating,gabriel-etal-2021-go},
repetition and redundancy
\cite{li-etal-2018-improving,suzuki-nagata-2017-cutting}, and struggle
to identify which content units are salient
\cite{tan-etal-2017-abstractive}. These phenomena are amplified when
generating long-form text, i.e.,~documents with multiple paragraphs
\cite{wiseman-etal-2017-challenges}, when dealing with non-linguistic
data (e.g.,~database tables), or very long input which is common when
summarizing multiple documents
\cite{liu-lapata-2019-hierarchical,topictempsum2019}, books
\cite{kryscinski2021booksum}, or dialogue
\cite{chen-etal-2022-summscreen,zhong-etal-2021-qmsum}.  An additional
challenge concerns the blackbox nature of deep learning systems which
hides the inherent complexity of modeling multiple interconnected
linguistic phenomena in text generation, and makes it difficult to
examine model decisions and attribute errors to specific
components. The lack of modularity further affects controllability as
these systems cannot be easily tailored to individual needs.

\begin{table*}[t!]
 \footnotesize
\center
\begin{tabular}{|p{15.3cm}|} \hline
\rowcolor{DarkBP}
\textcolor{black}{Q$_{1}$: Who built the Shelby Mustang from 1969 to 1970? \hfill A$_{1}$: Ford} \\\rowcolor{DarkBP}
\textcolor{black}{Q$_{2}$: During what years was the Shelby Mustang built by Shelby American? \hfill A$_{2}$: 1965 to 1968} \\\rowcolor{DarkBP}
\textcolor{black}{Q$_{3}$: In what year was the fifth generation of the Ford Mustang introduced? \hfill A${_3}$: 2005 } \\\rowcolor{DarkBP}
\textcolor{black}{Q$_{4}$: What was the Shelby Mustang revived as? \hfill A$_{4}$: a new high-performance model} \\ \hline\hline
\vspace{-0.1cm}
The Shelby Mustang is a high performance variant of the Ford Mustang which was built by 
Shelby American from 1965 to 1968, and from 1969 to 1970 by Ford.
Following the introduction of the fifth generation Ford Mustang in 2005, 
the Shelby nameplate was revived as a new high-performance model, 
this time designed and built by Ford.
\vspace{0.1cm}
\\\hline
\end{tabular}
\caption{\label{tab:qud:ex}  Question-answering (QA) blueprint for AQuaMuSe summary.
QA pairs were obtained from a
  state-of-the-art question generation and answer identification system \cite{alberti-etal-2019-synthetic}.}
\vspace{-0.5cm}
\end{table*}

Attempts to remedy some of these issues focus on changing the way
entities are represented
\cite{puduppully-etal-2019-data,iso-etal-2019-learning}, allowing the
decoder to skip low-confidence tokens to enhance faithful generation
\cite{tian2019sticking}, modeling graph connections between document
elements to better capture salience
\cite{tanjiwei2017,liu-lapata-2019-hierarchical}, encoding documents
hierarchically
\cite{celikyilmaz-etal-2018-deep,liu-lapata-2019-hierarchical,Rohde2021HierarchicalLF},
learning latent alignments between the input and the target text
\cite{xu-etal-2021-agggen} adopting sparse attention mechanisms
\cite{Child2019GeneratingLS,Beltagy2020LongformerTL}, and introducing
content selection \cite{gehrmann2018,dou-etal-2021-gsum} and planning
components
\cite{DBLP:journals/corr/abs-1809-00582,moryossef-etal-2019-step,narayan-etal-2021-planning,wiseman-etal-2018-learning}.

In this paper we also aim to render conditional generation more
modular via an intermediate, \mbox{plan-based} representation.  While
autoregressive models of language predict one token at a time, there
is evidence that in humans some degree of planning occurs at a higher
level than individual words \cite{Levelt:1993,Guhe:2007}. A long
tradition in natural language generation views planning as a central
component to identifying important content and structuring it
appropriately \cite{reiter-dale:00}, however, there is less agreement
on how plans should be represented. Common examples include discourse
trees \cite{mellish-etal-1998-experiments}, entity transitions
\cite{kibble-power-2004-optimizing,barzilay-lapata-2008-modeling},
sequences of propositions \cite{Karamanis:2004}, and schemas
\cite{McKeown:1985}.

Our work proposes a new conceptualization of text plans as a sequence
of \emph{question-answer} pairs. Specifically, we draw inspiration
from the ``Questions under Discussion'' (QUD) theory of discourse
structure which posits that one way of articulating the structure of a
text is to identify the questions and sub-questions that are raised
and answered by subsequent spans of text
\cite{Carlson:1983,Ginzburg:1994,vanKuppevelt:1995,Larsson:2002,roberts:2012:information,Reister:2019}. Theoretical
models of Questions under Discussion assume that discourse contains
\emph{implicit} questions for each of the assertions made which are
thereby turned into answers. These questions and answers can be
understood in terms of their use in moving a discourse forward to
achieve communicative goals. 
We propose to make QUDs \emph{explicit} by exploiting state-of-the-art
question generation technology
\cite{alberti-etal-2019-synthetic,lu-lu-2021-survey} and use them as
an intermediate representation layer for conditional generation,
i.e.,~a question-answering (QA) \emph{blueprint} operating as a proxy
for both content selection (i.e.,~what to say) and planning (i.e.,~in
what order).

Table~\ref{tab:qud:ex} illustrates a plan for generating a Wikipedia abstract 
from the AQuaMuSe dataset \cite{kulkarni2020aquamuse}. We enhance
existing datasets (e.g.,~for summarization) with similar blueprints
which we obtain automatically. We then convert input-output pairs into
input-blueprint-output tuples and propose to learn encoder-decoder
models from these augmented annotations. We develop three models which
vary in how they integrate blueprints in the generation process and
their ability to handle long outputs.
Aside from generating blueprints and their corresponding text in one
go, we propose a new architecture which iteratively plans and
generates a sentence at a time, conditioning on the input \emph{and}
the output sentences generated so far.
We do not generate a \emph{global} blueprint, rather our planning
process is \emph{incremental} and informed by generation, which we
argue affords greater control over the output and its
fluency. Moreover, the model is better equipped for long-form
generation, since it does not have to (autoregressively) decode the
blueprint and its summary in one go, avoiding the risk of
exceeding the maximum decoder length.

We instantiate our models with a Transformer \cite{NIPS2017_7181}
encoder-decoder architecture and perform experiments on summarization
datasets representing different information seeking tasks, application
domains, and user requirements.\footnote{Our models, training data and predictions will be made available at \url{https://github.com/google-research/language/tree/master/language/text-blueprint}.} In all cases, we empirically
demonstrate that blueprint models are more factual than alternatives
which do not resort to planning; we also observe that QA blueprints
are a better representation compared to plans based on entity chains
\cite{narayan-etal-2021-planning}, allowing tighter control of the
 output, and providing a comprehensive explanation for model
predictions (if the plan is erroneous, then the summary will be too).


\section{Related Work}
\label{sec:related-work}

\paragraph{Questions under Discussion} 

The Question Under Discussion (QUD)-based approach to discourse
structure assumes an open-ended inventory of possible questions and
sub-questions \cite{vanKuppevelt:1995}.  Recent efforts
\cite{de-kuthy-etal-2018-qud,westera-etal-2020-ted,Reister:2019} have
nevertheless shown it is possible to manually annotate documents with
QUDs, i.e.,~to formulate a question for every assertion expressed in a
text. \citet{de-kuthy-etal-2020-towards} even go as far as to
partially automate QUD annotation in German by automatically
generating all potentially relevant questions for a given
sentence. Related work \cite{ko-etal-2020-inquisitive} focuses on the
generation of inquisitive questions that reflect general text
understanding and free-form open-ended questions
\cite{Ko2021DiscourseCA}.  Our work builds upon QUD and related
discourse structure theories, however, we do not directly implement
any of them in particular. We adopt question answering as a good way
of spelling out the connection between the information structure of a
sentence and the discourse in which the sentence can function.


\paragraph{QA Pairs as a Proxy for Annotation Labels}
Question-answer pairs have been previously used as a proxy for
expressing semantic content. \mbox{QA-SRL}
\cite{he-etal-2015-question} 
is a representation
based on QA pairs which has been shown to capture the
vast majority of arguments and modifiers in PropBank
\cite{palmer-etal-2005-proposition} and NomBank
\cite{meyers-etal-2004-nombank}.  Instead of using a pre-defined role
lexicon, \mbox{QA-SRL} labels semantic roles with questions, whose
answers denote the argument bearing the role.  Follow-on work uses QA
pairs to represent discourse relations
\cite{pyatkin-etal-2020-qadiscourse} and to capture overlap
or redundancy at the propositional level
\cite{brook-weiss-etal-2021-qa}. We also employ QA pairs as an
abstraction of propositional content, however, we do not target
specific relation types, or make any linguistic assumptions about them
(e.g.,~discourse relations vs semantic roles).


\paragraph{Question-Answering in Summarization} 

QA pairs have been used for evaluating summaries
\cite{deutsch-roth-2021-understanding,eyal-etal-2019-question,durmus-etal-2020-feqa,wang-etal-2020-asking},
specifically as a means of estimating the information overlap between
a reference summary and a system-generated one. QA-based signals have
also been incorporated in the training of summarization models, using
reinforcement learning
\cite{arumae-liu-2018-reinforced,arumae-liu-2019-guiding,scialom-etal-2019-answers}
or as a way of identifying salient content in the input document
\cite{Deutsch2021QuestionBasedSS}. \citet{cao-wang-2022-hibrids}
introduce the task of hierarchical question-summary generation, where
a source document is condensed into \emph{multiple} summaries each answering
a different question. Questions are organized hierarchically into
broad questions and more specific sub-questions which are learned from
manual annotations. Our model outputs a QA-based plan and a
\emph{single} summary for a given document, although it is possible to
generate different summaries from different plans for the same
document. Our QA~pairs are obtained automatically and they are not
stuctured. 

\paragraph{Planning in Encoder-Decoder Models} 
Various recent efforts have developed planning modules in the context
of data-to-text generation. In most cases, the plans are specific to
the input which varies from tables and records to RDF tuples.  For
instance, \citet{DBLP:journals/corr/abs-1809-00582} learn a plan
corresponding to a sequence of records, and generate a summary
conditioned on it. \citet{narayan-etal-2020-stepwise} treat content
selection as a task similar to extractive summarization, they first
extract sentence plans and then verbalize them
one-by-one. \citet{moryossef-etal-2019-improving,moryossef-etal-2019-step}
propose a symbolic planning stage followed by a neural realization
stage.  Other work
\cite{puduppully-lapata-2021-data,Puduppully:ea:2022} advocates macro
planning, where document content is organized into a sequence of
paragraph plans which are verbalizations of tabular input. Our work is
closest to \citet{narayan-etal-2021-planning} who also target
summarization applications and learn an intermediate plan to guide
generation. We adopt a more elaborate plan representation based on QA
blueprints, and interface decoding with plan generation similarly to
\citet{narayan-etal-2020-stepwise}.


\section{Text Generation with Blueprints}
\label{sec:modeling-approach}

\begin{table*}[th!]
\scriptsize
\center
\begin{tabular}{| p{13cm} |c|c|c|} \hline
\multicolumn{1}{|c|}{{Overgenerated Question-Answer Pairs}} & RT & RH & CO \\ \hline
Q$_1$: What is a high performance variant of the Ford Mustang? \hfill A$_1$: The Shelby Mustang   & \cmark & \xmark & \\ 
Q$_2$: What is the high performance variant of the Ford Mustang called? \hfill A$_2$: Shelby   & \cmark & \xmark & \\ 
Q$_3$: What is a high performance variant of the Ford Mustang? \hfill A$_3$: Shelby Mustang   & \cmark & \xmark & \\ 
Q$_4$: What is a Shelby Mustang? \hfill A$_4$: a high performance variant   & \cmark & \xmark & \\ 
Q$_5$: The Shelby Mustang is a high performance variant of what? \hfill A$_5$: the Ford Mustang   & \cmark & \cmark  & \xmark  \\ 
Q$_6$: The Shelby Mustang is a high performance variant of what? \hfill A$_6$: Ford Mustang   & \cmark & \xmark & \\ 
Q$_7$: The Shelby Mustang is a high performance variant of what Ford model? \hfill A$_7$: Mustang   & \cmark & \xmark & \\ 
\hdashline
Q$_8$: Who built the Shelby Mustang from 1965 to 1968? \hfill A$_8$: Shelby American   & \cmark & \cmark & \xmark\\ 
\hdashline
Q$_9$: During what years was the Shelby Mustang built by Shelby American? \hfill A$_9$: 1965 to 1968 & \cmark & \cmark & \cmark  \\ 
Q$_{10}$: In what year did Ford take over production of the Shelby Mustang? \hfill A$_{10}$: 1969   & \cmark & \xmark & \\ 
\hdashline
Q$_{11}$: What was the final year that Shelby American built the Mustang? \hfill A$_{11}$: 1970   & \xmark & & \\ 
Q$_{12}$: Who built the Shelby Mustang from 1969 to 1970? \hfill A$_{12}$: Ford   & \cmark & \cmark & \cmark \\ 
\hdashline
Q$_{13}$: What event in 2005 led to the revival of the Shelby Mustang? \hfill A$_{13}$: the introduction   & \xmark & & \\ 
Q$_{14}$: What generation of Mustang was introduced in 2005? \hfill A$_{14}$: the fifth generation   & \cmark & \xmark & \\ 
Q$_{15}$: What generation of Mustang was introduced in 2005? \hfill A$_{15}$: fifth   & \cmark & \xmark & \\ 
Q$_{16}$: In what year was the fifth generation of the Ford Mustang introduced? \hfill A$_{16}$: 2005   & \cmark & \cmark & \cmark \\ 
\hdashline
Q$_{17}$: What name was brought back for the 2005 Ford Mustang? \hfill A$_{17}$: the Shelby nameplate   & \cmark & \xmark & \\ 
Q$_{18}$: What was the Shelby Mustang revived as? \hfill A$_{18}$: a new high-performance model   & \cmark  & \cmark & \cmark \\ 
\hline \hline
\multicolumn{4}{|p{15.5cm}|}{[The Shelby Mustang is a high performance variant of the Ford Mustang]$_{P_1}$ which [was built by Shelby American]$_{P_2}$ [from 1965 to 1968,]$_{P_3}$ and [from 1969 to 1970 by Ford.]$_{P_4}$ [Following the introduction of the fifth generation Ford Mustang in 2005,]$_{P_5}$ [the Shelby nameplate was revived as a new high-performance model, this time designed and built by Ford.]$_{P_6}$}
\\
\hline
\end{tabular}
\vspace{-0.2cm}
\caption{\label{tab:qgen-filter-ex} Generation of QA pairs for 
  summary in Figure~\ref{tab:qud:ex} and blueprint annotation. We split the summary into propositions~$P$ and select no more
  than one QA pair per proposition. RT, RH, and CO are shorthands for
  Round Trip, Rheme, and Coverage. Questions that pass/fail each
  filter are marked with \cmark/\xmark.}
\vspace{-0.5cm}
\end{table*}

\subsection{Problem Formulation}

Let~$d$ denote the input to the model which could be a document (or
multiple documents), a dialogue history, or even database tables. The model will learn to generate  blueprint~$b$ for output~$s$
(e.g.,~a summary) and the output itself. The blueprint~$b$ is an ordered
set of question-answer pairs $\{(q_1, a_1), (q_2, a_2), \ldots, (q_m,
a_m)\}$. Unsurprisingly, such blueprints are not naturally occurring
in existing datasets which typically consist of~\mbox{$(d, s)$}
pairs. In the following we explain how we automatically augment
training examples $(d, s)$ into tuples~$(d, b, s)$ with blueprints
(Section~\ref{subsec:annotation}) and then describe how we devise
blueprint models based on them~(Section~\ref{subsec:models}).


\subsection{Blueprint Annotation}
\label{subsec:annotation}
We first explain how question-answer pairs are automatically
(over-)generated for output~$s$, and subsequently filtered to create
blueprint~$b$. We illustrate the different filtering stages via the
example in Table~\ref{tab:qgen-filter-ex}.

\paragraph{Question-Answer Generation}


We generate QA pairs following an approach similar to
\citet{honovich2021q2, honovich-etal-2022-true}. We convert the SQuAD
reading comprehension dataset ~\cite{rajpurkar2018know} to a question
generation dataset by concatenating the answer and context (with
separators) and fine-tuning a sequence-to-sequence transformer model
to predict the question.  Specifically, we fine-tune the T5-11B
checkpoint from \citet{Raffel:ea:2019}; questions are decoded with a
beam size of~4. During training, answer candidates are the answers
provided in the SQuAD annotation.  At inference time, answer
candidates (i.e.,~base noun phrases and named entities) are identified
in the output~$s$ using SpaCy\footnote{\url{https://spacy.io/}} and
questions are generated with the SQuAD trained system. This procedure
yields a large list of QA pairs (see in Table~\ref{tab:qgen-filter-ex}
the questions generated for the summary at the bottom), which
we reduce using the filtering explained below.


\paragraph{Question-Answer Blueprints}

Initially, we apply a {\em Round-trip Consistency} check
\cite{alberti-etal-2019-synthetic} which discards questions if they
yield answers different from those used to generate them. In
Table~\ref{tab:qgen-filter-ex}, Q$_{11}$ is discarded as the answer it
is paired with is wrong (\textsl{1968} was the final year that Shelby
American built the Mustang, not \textsl{1970}). The same is the case
for Q$_{13}$, where the answer to the question ought to have been
\textsl{the introduction of the of the fifth generation Ford Mustang}.

To decrease the number of QA pairs further, we chunk the text (bottom
block in Table~\ref{tab:qgen-filter-ex}) into \textit{propositions}
--- a proposition is a sub-sentential unit which represents a single
claim or fact \cite{stanovsky-etal-2018-supervised,proposition-sum21},
We use propositions instead of sentences since the latter can be too
long and contain multiple facts.  We split text into propositions
based on punctuation (period, comma, and semicolon), coordination
(e.g.,~\textsl{and}, \textsl{but}), relative pronouns
(e.g.,~\textsl{that, who}), and prepositions (e.g.,~\textsl{at,
  by}). Following this simple approach, the summary in
Table~\ref{tab:qgen-filter-ex} is split into six propositions, shown
within square brackets.  We next match each proposition to a
\emph{single} QA pair heuristically, following a two-stage approach.

We first find the question whose answer is at the rightmost position
within a proposition.  If there are multiple such questions, we select
the one with the longest answer. This first stage, which we call {\em
  Rheme}, is motivated by the theme-rheme structure
\cite{vallduvi1998rheme} of natural language sentences: already known
information (i.e., the theme) is usually placed first while new
information (i.e., the rheme) is placed later in a sentence or phrase
\cite{kruijff2003discourse}. Following this idea, Rheme selection
prioritizes new-information seeking questions. As can be seen in
Table~\ref{tab:qgen-filter-ex},  it eliminates several
questions (e.g., Q$_{1}$--Q$_{4}$) as their answers are not the right
most element in the obtained propositions. Questions Q$_{5}$ and
Q$_{6}$ are identical, however we retain Q$_{5}$ as it yields the
longest answer.


The second stage, which we call {\em Coverage}, prioritizes the
selection of informative QA pairs by selecting non-overlapping
ones. Specifically, we first convert~$s$ to a bag of tokens and select
the QA pair with the highest lexical overlap. We then remove the
overlapping tokens from~$s$, and repeat this greedy selection process
until the bag is empty or the overlap is
zero. Table~\ref{tab:qgen-filter-ex} shows how Coverage further
eliminates QA pairs Q$_{5}$ and Q$_{8}$. The remaining four QA pairs
constitute the final blueprint~$b$. Rather than defaulting to a random
order, we sort these based on the location of the answer spans
in~$s$ (see the final order in Table~\ref{tab:qud:ex}).

\subsection{Blueprint Models}
\label{subsec:models}

We devised three seq-to-seq models, which differ in the way the
output and its blueprint are
generated.


\paragraph{End-to-End Model}


A straightforward approach would be to take~$d$ as input and learn to
first predict blueprint~$b$ as~$p(b|d)$, and then generate output~$s$
as~$p(s|b)$. However, this approach crucially relies on the blueprint
being accurate and capturing all required information, which might be
overly optimistic, given that blueprints (for training) are generated
automatically. Moreover, pipeline architectures are known to suffer
from error propagation, which in our case would undoubtedly affect
generation performance, the final stage of the pipeline.

Rather than modeling the blueprint and output generation stages
separately, we  train an encoder-decoder model to encode~$d$ and
generate~$b;s$ (i.e.,~the concatenation of the blueprint and output
sequence) in one go. Essentially, the decoder first predicts
blueprint~$b$ and then continues to generate output~$s$, using
both~$b$ and~$d$. We prefix~$b$ and~$s$ with special markers
``\emph{Plan:}'' and ``\emph{Summary:}'', respectively. In particular,
we predict~$b$ as $a_1;q_1;\ldots;a_m;q_m$, namely a (concatenated)
sequence of answer-question pairs.\footnote{Predicting~$b$ as
  $q_1;a_1;\ldots;q_m;a_m$ is more natural, however, it led to inferior performance. 
 See the ablation experiments in Section~\ref{sec:ablations}.} The model is trained with the standard maximum-likelihood objective
to generate the augmented target~$b;s$. Interestingly, in this
end-to-end model the blueprint functions as a \emph{macro-plan},
i.e.,~a \emph{global} sketch of the content and organization of the
output.

\paragraph{Multi-task Model}


It is generally challenging for encoder-decoder models to generate
long output sequences
\cite{ko-li-2020-assessing,tan-etal-2021-progressive}. The end-to-end
model sketched above further amplifies this problem because it
ultimately aims to generate sequence $b;s$ rather than just~$s$,
increasing the sequence length by~220\% (see Table~\ref{tab:dataset}).

To mitigate this problem, we propose a multi-task model optimized to
perform two separate tasks.  Let~$a$ and~$q$ denote an ordered
sequence of answers ($a_1, \ldots, a_m$) and corresponding questions
($q_1, \ldots, q_m$), in blueprint~$b$. The model is trained to
generate (a)~the answer plan concatenated with output sequence~$a;s$,
and (b)~the answer plan concatenated with  questions~$a;q$. In
particular, we train a single encoder-decoder model to encode
input~$d$, while the decoder first predicts answer plan~$a$
(as~$p(a|d)$) and then continues to generate output~$s$
(as~$p(s|a,d)$) or corresponding questions~$q$ (as~$p(q|a,d)$),
depending on the task. We prefix $a$, $q$, and $s$ with special
markers ``\emph{Plan:}'', ``\emph{Questions:}'', and
``\emph{Summary:}'', respectively. We further prefix input~$d$ with
``\emph{Generate Summary:}'' or ``\emph{Generate Questions:}'' to
instruct our model to generate output~$s$ or questions~$q$,
respectively. We sample data points from these two tasks with equal
probability and train the model with the standard maximum-likelihood
objective.

During inference, we use a two-step process to generate output~$s'$
and its blueprint~$b'$ for input~$d$. We first prefix~$d$ with
``\emph{Generate Summary:}'' and generate~$a';s'$, i.e.,~answer
plan~$a'$ followed by output sequence $s'$. We then prefix~$d$ with
``\emph{Generate Questions:}'', prompt our decoder with the predicted
answer plan $a'$ and generate corresponding questions~$q'$ for
blueprint~$b'$. The multi-task model alleviates the length issue
discussed above by learning to generate $a;s$ instead of
$b;s$. However, this comes at the expense of generation quality, since
the model now conditions on the answers only, not question-answer
pairs. As such, it can be viewed as an extension of FROST
\cite{narayan-etal-2021-planning} with the plan being a sequence of
\emph{answer spans} rather than \emph{entity chains}. This model also
creates a macro-plan of the output, however, less detailed compared to
the end-to-end model. 

\paragraph{Iterative Model} 


Rather than predicting a global plan (i.e., answer plan~$a$ or
blueprint~$b$) prior to generating output~$s$, we employ an
incremental approach which interleaves planning with text
generation. Let output~$s$ consist of $n$~sentences $\{s_1,s_2,
\ldots, s_n\}$; then, the corresponding blueprint~$b$ can be
represented as $\{b_1,b_2, \ldots, b_n\}$, where $b_i: \{(a_{j+1}^i,
q_{j+1}^i), \ldots, (a_{j+k}^i, q_{j+k}^i)\}$ consists of $k$
question-answer pairs for sentence~$s_i$. We train our model to
iteratively plan and generate one sentence at a time, conditioning on
the input and the output sentences generated so far. In particular, we
train an encoder-decoder model where the encoder first encodes
input~$d$, while the decoder takes summary $\{s_1, \ldots, s_i\}$
generated so far as a prompt and generates blueprint $b_{i+1}$ for
next sentence $s_{i+1}$, followed by sentence~$s_{i+1}$ itself.

The iterative model is trained on quadruples $\{(d,
\phi, b_1, s_1), \ldots, \allowbreak (d, s_{1,i}, b_{i+1}, s_{i+1}),
\ldots, \allowbreak (d, s_{1,n-1}, b_n, s_n), \allowbreak (d, s,
\allowbreak b_{end}, s_{end})\}$, where~$\phi$ is an empty context
placeholder used to predict the first blueprint $b_1$ and corresponding first
sentence~$s_1$,  $(n+1)$ is the blueprint length, and $s_{1,i}=\{s_1,
\ldots, s_i\}$ are the output sentences generated so far;  $b_{end}$
and $s_{end}$ are special tokens marking the end of the output
prediction. We prefix $s_{1,i}$, $b_i$, and $s_i$ with special markers
``\emph{Context:}'', ``\emph{Plan:}'', and ``\emph{Next Sentence:}'',
respectively. We train the model with the standard maximum-likelihood
objective to predict $s_{1,i};b_i;s_i$, however, we do not compute
the loss for predicting  context~$s_{1,i}$ to avoid over-optimizing for
sentences that appear at the beginning of the output.


The iterative approach does not create a global macro plan. Rather, it
learns \emph{micro} content plans and verbalizes them one-by-one,
conditioning on previously generated sentences but not on previously
generated QA pairs. Athough it does not have a global document
view like the end-to-end model, the iterative decoder cannot exceed
the output sequence length as it plans and predicts one sentence at a
time as $b_i;s_i$, instead of generating~$b;s$ in one go. And unlike
the multi-task model, each sentence~$s_i$ is generated by conditioning
on the full blueprint~$b_i$ (consisting of questions \emph{and}
answers).


\section{Experimental Setup}
\label{sec:exp-setup}

\begin{table}[t!]
\footnotesize
\centering
\begin{tabular}{@{}l@{~}|l@{~}|l@{~}|l@{}} \hline
&  \multicolumn{1}{c|}{{AQuM}} &
  \multicolumn{1}{c|}{{WCSum}} &
  \multicolumn{1}{c}{{SS-FD}} \\   \hline
\# queries & 8,162 & \multicolumn{1}{c|}{---} & \multicolumn{1}{c}{---}  \\
\# examples & & & \\
\hspace{0.3cm} train & 6,599 & 165,000 & 3,673 \\
\hspace{0.3cm} dev & 714 & 8,723 & 338 \\
\hspace{0.3cm} test & 849 & 9,166 & 337 \\
source & & & \\
\hspace{0.3cm} \# docs & 6.46 & 135.56 & 1.00 \\
\hspace{0.3cm} \# words & 12,986.88 & 7455.75 & 8051.74 \\
\hspace{0.3cm} \# sentences & 339.62 & 307.80 & 804.01 \\
\hspace{0.3cm} \# words/doc & 2,008.38 & 52.02 & 8051.74 \\
target (original) & & & \\
\hspace{0.3cm} \# words & 114.07 & 115.61 & 126.73\\
\hspace{0.3cm} \# sentences & 3.65 & 4.74 & 5.26  \\
\hspace{0.3cm} novel unigrams & 0.02 & 0.13 & 0.17 \\
\hspace{0.3cm} novel bigrams & 0.13 & 0.54 & 0.66 \\
\hspace{0.3cm} novel trigrams & 0.24 & 0.78 & 0.92 \\
\hspace{0.3cm} novel 4-grams & 0.31 & 0.86 & 0.98 \\
target (+blueprint) &  & &\\
\hspace{0.3cm} \# QA-Pairs &  8.16 & 9.56 &  28.10 \\
\hspace{0.3cm} \# words& 272.68 & 291.28 & 597.90 \\\hline
\end{tabular}
\vspace{-0.1cm}
\caption{\label{tab:dataset} Summary statistics for the datasets used
  in this work (AQuM, WCSum, and SS-FD are shorthands for AQuaMuse,
  WikiCatSum and ScreenSumm-FD, respectively).  We report on the
  number of queries, size of training, development, and test set, and
  average source and target length (in terms of documents, words,
  sentences, and words per document). We quantify the
  abstractiveness of the target by measuring the proportion of 
  $n$-grams unseen in the source. We also report statistics on the
  target length augmented with the blueprint (number of QA pairs and
  words in total).}
  \vspace{-0.5cm}
\end{table}





\subsection{Datasets}
We evaluated our model on benchmarks representative of long-form
question answering and summarization.  Our datasets vary in terms of
the input given to the generation model (e.g., multiple documents or
one, web pages, or dialogue transcripts), the user's information need
(e.g., answering a question or aggregating information), and summary
style (e.g., genuinely abstractive vs extractive). Common features
among them are very long inputs and multi-sentence output
summaries. We summarize various dataset statistics in
Table~\ref{tab:dataset}.

\paragraph{AQuaMuSe} \cite{kulkarni2020aquamuse,aquamuse21} is a query-focused
multi-document summarization dataset; it was created with the intent
of simulating how a search engine might synthesize documents of high
relevance to a user query. It consists of Google Natural Questions
\cite{kwiatkowski-etal-2019-natural} paired with web documents
extracted from Common Crawl and long-form answers from Wikipedia. 
We approach this task as a generative QA problem where we take the
query and associated web documents and generate a long-form answer to
the query. We  work on the split from
\newcite{aquamuse21}; 
on average, each instance has 6.46 web documents (2,008 tokens per
document), leading to very long input (12,987 tokens).

\paragraph{WikiCatSum} \cite{topictempsum2019} is a topic-focused multi-document
summarization dataset
where the goal is to generate Wikipedia abstracts (i.e.,~lead article
sections) from a large set of webpages related to an entity or a
topic.
It focuses on three entities, namely Films (59,973~instances),
Companies (62,545~instances), and Animals (60,816~instances). In
experiments, we collate the different data subsets into one which we
refer to collectively as WikiCatSum. The input webpages are truncated
to the first 800~tokens.

%

\paragraph{SummScreen-FD} \cite{chen-etal-2022-summscreen} is a
recently released dialogue summarization dataset. It contains
transcripts of TV episodes (e.g., Game of Thrones, CSI Las Vegas) and
corresponding (community authored) summaries.  The original dataset is
divided into two complementary subsets; we use the ForeverDreaming
(FD) subset released as part of the \textsc{Scrolls} benchmark
\cite{Scrolls:2022} which incorporates episodes from 88~different
shows. SummScreen-FD is a challenging testbed for several
reasons. Plot details are often expressed indirectly in conversations
between characters and are scattered across the entire transcript. The
summarization task is highly compressive, a transcript the size of a
book (on average 8,000 tokens; see Table~\ref{tab:dataset}) is
condensed into a few sentences, and the evaluation of such summaries
comes with its own challenges (e.g., it is not realistic to expect
humans to read the transcript to be able to assess their quality).

We further analyze the characteristics of these datasets in
Table~\ref{tab:dataset}. Long-form answers in AQuaMuSe are mostly
extractive with only 2\%, 13\%, 24\%, and 31\% novel unigrams,
bigrams, trigrams, and 4-grams, respectively. In comparison, summaries
in WikiCatSum and SummScreen-FD are more abstractive; WikiCatSum
abstracts have 13\% novel unigrams, 54\% bigrams, 78\% trigrams, and
86\% 4-grams, whereas in SummScreen-FD summaries 17\% unigrams, 66\%
bigrams, 92\% trigrams, and 98\% \mbox{4-grams} were not seen in the
training. Interestingly, SummScreen-FD summaries have far more
propositions than AQuaMuSe or WikiCatSum targets, leading to a much
higher number for QA pairs in their blueprints (28.10 vs 8.16 or
9.56). This in turn makes the generation task for end-to-end models
very challenging.  The average summary length together with the
blueprint annotations (i.e.,~$b;s$) for SummScreen-FD is almost twice
the size of WikiCatSum and AQuaMuSe (597.90 vs 291.28 and 272.68).
The majority of questions in AQuaMuSe and WikiCatSum are
\textsl{what} questions (76.0\% and 74.2\%, respectively), followed by
\textsl{who}, \textsl{where}, \textsl{when}, and \textsl{how}
questions. For SummScreen-FD, {\em what} and {\em who} questions are
most popular (50.1\% and 42.9\%, respectively).

\subsection{Comparison Systems}

All our experiments used  \longtv\ \cite{Guo:ea:2022}, an extension
of the original \tv\ \emph{encoder} \cite{Raffel:ea:2019} with global-local attention
sparsity patterns to handle long inputs. We compared a vanilla  \longtv\footnote{We
  used the publicly released checkpoints from
  \url{https://github.com/google-research/longt5}} model (xl,
3B parameters) fine-tuned on our datasets (with a maximum input sequence length of
4,096 tokens and a maximum output length of 512 tokens) against several blueprint variants. 
These include
an  end-to-end \longtv\ model (\etoe)  which first decodes blueprint~$b$
and then continues to decode output~$s$; a \longtv\ multitask model
(\aplan) which jointly learns to predict the answer plan followed by
either the output~$s$ or the questions in~$b$; and a
\longtv\ iterative model (\iterative) which plans and generates one
sentence at a time. 

In addition, we implemented a two-stage model
(\twostage) which first creates blueprint~$b$ given input~$d$ and then
generates output~$s$ given $b$ and $d$ as input.  Finally, we also
fine-tuned \tv\ (xl,
3B parameters) on our datasets with a maximum input sequence length
of 1,024 tokens and a maximum output length of~256 tokens, as a
baseline. We present these comparisons in
Table~\ref{table:merged-results} together with the performance of
various state-of-the-art systems. 

We finetuned all our models with a leaning rate of 0.001 and a batch size of 128, for 50K steps. We select best checkpoints using average Rouge performance on validation sets. During inference, we use beam search with size 5 and alpha 0.8.


\section{Automatic Evaluation} 

In this section we present experimental results using automatic
evaluation metrics which assess overall summary (and blueprint)
quality. Moreover, we quantify the extent to which automatically
generated output is grounded to the blueprint and faithful to the
input document/s.

\subsection{Metrics}

\paragraph{Summary and Blueprint Quality} We evaluate summary quality
automatically using (summary-level) Rouge F1
\cite{lin-hovy-2003-automatic}. We report only
RougeLSum\footnote{RougeLSum is very similar to ROUGE-L; while the
  latter is calculated on the summary as a whole, RougeLSum interprets
  newlines as sentence boundaries.} in Table~\ref{table:merged-results}
for the sake of brevity.
We also use RougeLSum to evaluate the quality of the automatically
generated blueprint, i.e., the QA pairs and their order against the
reference blueprint.

\paragraph{Informativeness and Grounding} We evaluate informativeness
using QA-based metrics. Specifically, following the reading
comprehension literature
\cite{rajpurkar-etal-2016-squad,rajpurkar-etal-2018-know}, we quantify
the extent to which the generated text can answer all questions from
its reference ({\em Informativeness}) and predicted blueprint ({\em
  Grounding}). Following \newcite{stelmakh-etal-2022-asqa}, we use a RoBERTa model \cite{roberta}
finetuned on SQuAD-V2 for question-answering in both cases.\footnote{This is a high
  performing model reaching   86.8\% exact-match accuracy and~89.8\%
  F1 on SQuAD.} Given generated text~$s'$ and question-answer
pair~$(q_i, a_i)$ from the (reference or predicted) blueprint, we
apply our question-answering model to~$s'$ to predict answer~$a'_i$ to
question~$q_i$. We then compute the token-level F1 score between
predicted answer~$a'_i$ and ground truth answer~$a_i$, and report the
average.

\paragraph{Faithfulness}  

Hallucinations are a widely known issue with neural abstractive
summarization
\cite{song-etal-2018-structure,maynez-etal-2020-faithfulness,kryscinski-etal-2020-evaluating,gabriel-etal-2021-go},
especially when a sentence combines content from multiple sources
\cite{lebanoff-etal-2019-analyzing}.

Following previous work
\cite{maynez-etal-2020-faithfulness,falke-etal-2019-ranking,narayan-etal-2022-well,honovich-etal-2022-true,dusek-kasner-2020-evaluating},
we quantify the extent to which generated summaries are faithful to
their input using textual entailment. We resort to textual entailment
for two reasons; firstly, it is a relatively intuitive metric, all
information in a summary should be entailed by the source or at least
not conflict with it; secondly recent studies
\cite{maynez-etal-2020-faithfulness,Fischer:ea:2022}, have shown that
it correlates with human judgments of faithfulness across
summarization datasets and tasks.

Following \newcite{honovich-etal-2022-true}, we trained an entailment
model by fine-tuning T5-11B \cite{Raffel:ea:2019} on the Adversarial
NLI dataset (ANLI; \citealt{nie-etal-2020-adversarial}).
For each sentence (hypothesis) in the summary, we compute its
entailment probability given the input (premise) and report the
average across all sentences to obtain an overall score
\cite{maynez-etal-2020-faithfulness}. 

More formally, let~$E$ denote a textual entailment model that
predicts $E(a,b)$, namely that text~$b$ is entailed by text~$a$.  The
faithfulness score~$F$ of summary~$s$ containing sentences
$s_1,\dots,s_2$ with respect to input~$D$ is computed as: 
\[
F(s) = \frac{1}{n}\sum^{n}\limits_{i=1}E(D,s_i)
\]
where $n$~is the number of sentences in the summary.
If the input is longer than the T5 maximum encode length, we split it,
calculate the entailment probability per split, and take the
maximum. We convert probabilities to binary labels using a threshold
(\mbox{1 if $>$ 0.5}, and 0, otherwise).

We further validated our
  ANLI entailment scores against human judgements of faithfulness elicited as part of SummEval \citep{fabbri-etal-2021-summeval}, a recently released dataset for assessing automated summarization metrics. Our entailment predictions correlate well with human ratings, achieving a Spearman's rank correlation  of~$\rho=0.774$.

\subsection{Results}
\label{sec:results}

\paragraph{Why  \longtv\  for Blueprint Models}

All the tasks we are dealing with require modeling input of highly
complex nature  which is often very long (see
Table~\ref{tab:dataset}). 
Our results in Table~\ref{table:merged-results} (see Rouge/summary
column) demonstrate that \tv\ models always fall behind \longtv,
underscoring the importance of sparse attention mechanisms for
modeling long inputs. In fact, \longtv\ sets a new state of the art on
AQuaMuSe and SummScreen-FD. On WikiCatSum, it is slightly worse than
\textsc{Reflect} \cite{reflect}, an extract-then-abstract model which
has a dedicated content selection module. Similar content selection
techniques could also benefit \longtv, however, we leave this to
future work. We henceforth use \longtv\ as a base model for finetuning
our blueprint models.


\begin{table}[t!]
\center
\resizebox{\linewidth}{!}{
{\setlength{\tabcolsep}{0pt}%
\begin{tabular}{@{}l l| l l |l l |l @{}} \hline
& & \multicolumn{2}{c|}{RougeLSum} &  \multicolumn{2}{c|}{QA-F1} & \multicolumn{1}{c}{ANLI}  \\ 
& \multicolumn{1}{c|}{Models} & \multicolumn{1}{c}{~summ.~~} & \multicolumn{1}{c|}{bluepr.} & \multicolumn{1}{c}{~inform.~~} & \multicolumn{1}{c|}{ground.} & \multicolumn{1}{c}{~entail.} \\ \hline 
&\textsc{HiBERT}* & ~27.11 & \hspace*{1.2ex}--- & \hspace*{1.2ex}--- & \hspace*{1.2ex}--- & \multicolumn{1}{c}{---} \\
& TextRank* & ~31.07 & \hspace*{1.2ex}--- & \hspace*{1.2ex}--- & \hspace*{1.2ex}--- & \multicolumn{1}{c}{---} \\ 
&\textsc{SiBERT}* & ~30.79 & \hspace*{1.2ex}--- & \hspace*{1.2ex}--- & \hspace*{1.2ex}--- & \multicolumn{1}{c}{---} \\ 
&\tv & ~41.53 & \hspace*{1.2ex}--- & ~19.77 & \hspace*{1.2ex}--- & ~70.01  \\
& \longtv & ~\textbf{61.43}  & \hspace*{1.2ex}--- & ~\textbf{39.22} & \hspace*{1.2ex}--- & ~82.52 \\ 
& \twostage & ~46.34 & 36.61 & ~24.10 & 64.22 & ~61.40  \\
  &\cellcolor{DarkBP}\textcolor{black}{\etoe} & ~58.96$^{\dagger}$ & 54.16$^{\dagger}$ & ~34.59 & \textbf{69.53} & ~83.64$^{\dagger}$  \\
&\cellcolor{DarkBP}\textcolor{black}{\aplan} & ~59.24$^{\dagger}$ & \textbf{54.88} & ~37.87$^{\dagger}$ & 52.01 & ~83.37 \\
\raisebox{5ex}[0pt]{\begin{sideways}AQuaMuse\end{sideways}}& \cellcolor{DarkBP}\textcolor{black}{\iterative} & ~56.48 &  52.92$^{\dagger}$ & ~36.04 & 69.37$^{\dagger}$ & ~\textbf{84.77} \\\hline
&\textsc{Bart} & ~39.73 & \hspace*{1.2ex}--- & \hspace*{1.2ex}--- & \hspace*{1.2ex}--- &\multicolumn{1}{c}{---}  \\  
&\textsc{Reflect} &  ~42.17 &  \hspace*{1.2ex}--- & \hspace*{1.2ex}--- & \hspace*{1.2ex}--- & \multicolumn{1}{c}{---}  \\  
&\tv & ~40.17 &  \hspace*{1.2ex}--- & ~18.79 & \hspace*{1.2ex}--- & ~46.21 \\
&\longtv &  ~41.00 & \hspace*{1.2ex}---  & ~\textbf{22.97}  & \hspace*{1.2ex}--- & ~56.11  \\ 
& \twostage & ~37.84 & 24.27 & ~17.78 & 79.48 & ~49.68 \\
&\cellcolor{DarkBP}\textcolor{black}{\etoe} & ~39.78 & 45.40 &  ~19.22 & \textbf{82.88}  & ~60.56\\
&\cellcolor{DarkBP}\textcolor{black}{\aplan} & ~\textbf{42.34}  & \textbf{46.95} & ~20.55$^{\dagger}$ & 62.14 & ~52.09 \\ 
\raisebox{3.6ex}[0pt]{\begin{sideways}WikiCatSum\end{sideways}} & \cellcolor{DarkBP}\textcolor{black}{\iterative} & ~39.53 & 46.35$^{\dagger}$ & ~18.75 & 80.35$^{\dagger}$  & ~\textbf{64.37} \\\hline
& \textsc{r2t-Bart} & ~23.30 & \hspace*{1.2ex}--- & \hspace*{1.2ex}--- & \hspace*{1.2ex}--- & \multicolumn{1}{c}{---}  \\  
& \tv & ~25.84 & \hspace*{1.2ex}--- & ~3.31 & \hspace*{1.2ex}--- & \hspace{1.4ex}0.78  \\
& \longtv & ~31.22$^{\dagger}$ & \hspace*{1.2ex}--- & ~\textbf{7.59} & \hspace*{1.2ex}--- & \hspace{1.4ex}8.40 \\ 
& \twostage & ~19.07 & 21.63 & ~2.70 & 54.22 & \hspace{1.4ex}1.24 \\
&\cellcolor{DarkBP}\textcolor{black}{\etoe} & ~28.22 & 34.49  & ~6.13 & 54.53 & \hspace{1.4ex}8.53\\
&\cellcolor{DarkBP}\textcolor{black}{\aplan} & ~\textbf{31.88} &
35.23 & ~7.06$^{\dagger}$ & 21.15 & \hspace{1.4ex}9.57 \\
\raisebox{2ex}[0pt]{\begin{sideways}SummScreen\end{sideways}}&\cellcolor{DarkBP}\textcolor{black}{\iterative}    & ~29.86$^{\dagger}$ & \textbf{39.36}  & ~7.55$^{\dagger}$ & \textbf{70.83} & \textbf{20.84}\\ 
\hline
\end{tabular}}
}
\caption{Results on AQuaMuSe, WikiCatSum, and SummScreen-FD test
  sets. Baseline and earlier SOTA models are presented in the top
  block and all blueprint models are shown in the bottom block. Models
  marked with * generate extractive summaries. \textsc{HiBERT},
  TextRank, and \textsc{SiBERT} results on AQuaMuSe are taken from
  \newcite{aquamuse21}.  \textsc{Bart} and \textsc{Reflect}
  (extract-then-abstract) results are taken from
  \newcite{reflect}. Hybrid \textsc{r2t-Bart} (content selection +
  generation) results are taken from
  \newcite{chen-etal-2022-summscreen}.  Best results for each task are
  \textbf{boldfaced}. Scores that are \uline{not} significantly different
  (using paired bootstrap resampling; \mbox{$p < 0.05$}) from the best
  score in each column are marked with a dagger
  (${\dagger}$). \label{table:merged-results} }
\end{table}

\paragraph{Blueprint Models and Rouge}

Compared to \longtv, blueprint variants slightly underperform on
AQuaMuse, but score better on WikiCatSum and SummScreen-FD (see
\aplan\ model). All differences between \longtv and blueprint models
are statistically significant using paired bootstrap resampling;
\mbox{p < 0.05}). For a fair comparison, we always use a maximum
decoder length of 512 tokens. With the exception of AQuaMuse,
\etoe\ is inferior to other blueprint models, which is not surprising
since it has to generate much longer text (recall it predicts $b;s$
rather than simply~$s$). Overall, \aplan\ is significantly better than
other blueprint models on WikiCatSum but on par with \iterative\ on
SummScreen-FD.

Similar patterns emerge when evaluating the predicted blueprints
against reference QA pairs, with \iterative\ significantly
outperforming the other two variants on SummScreen-FD. This could be
due to the fact that SummScreen-FD summaries have far more
propositions than AQuaMuSe or WikiCatSum targets; it is better to
predict them one sentence at a time, rather than all together. With
regard to WikiCatSum, the difference between \aplan\ and
\iterative\ is not significant (although \aplan\ has a slight numerical
advantage) and both systems are significantly better than \etoe. On
AQuaMuSe \aplan\ is significantly better than \etoe\ and \iterative.

Note that all \twostage\ models are significantly worse in comparison
to blueprint variants, when evaluating either their blueprints or
summaries (in terms of Rouge). While our models learn to optimize
blueprints and summaries \emph{together}, \twostage\ models are faced
with the harder task of predicting the blueprint solely based on the
input (text-to-data). Since the blueprints learned by the first stage
are of poor quality, the summaries generated in the second stage are
also inferior.


\paragraph{Blueprint Models and Informativeness}

Our blueprint annotation of reference summaries naturally provides a
more principled alternative to Rouge. We can now use QA pairs in
\emph{reference} blueprints to evaluate the informativeness of
\emph{predicted} summaries. Results follow a pattern overall similar
to Rouge, however, this approach reveals the complexity of the
different generation tasks better than Rouge. While we were able to
achieve reasonably high Rouge across datasets, we are far from
generating  informative summaries. On SummScreen-FD, in
particular, we achieve a maximum Rouge score of 31.88, but are able to
answer correctly only~7.59\% of reference questions using the
predicted summaries.

Across datasets, \longtv\ performs on par with \aplan, the
difference between the two models is not statistically significant,
and the same is true of \iterative\ on SummScreen.


\begin{table}[t]
\scriptsize
\resizebox{\linewidth}{!}{
  \begin{tabular}{@{}p{8cm}@{}} \hline
    \multicolumn{1}{c}{Reference Summary} \\ \hline
    Grissom and Catherine investigate the death of a man found in a dumpster, who is found to
  have had a severe eating disorder.  Meanwhile, Nick and Sara
  investigate the death of a couple at the brink of a bitter divorce,
  in which evidence seems to point to the couple's
  dog.\\\hline
  \multicolumn{1}{c}{\textsc{r2t-Bart}} \\\hline
  \textcolor{rred}{\textbf{Catherine Catherine}}, \textcolor{rred}{\textbf{Sara}} and Grissom investigate the death of a man who was found dead in a garbage bin. \textcolor{rred}{\textbf{The victim's wife, Lori Tinsley}}, was a poker player. \textcolor{rred}{\textbf{The man's brother, Jesse Menyel, was also found dead.}} The case is complicated by the fact that the victim had a gun in his apartment. Meanwhile, \textcolor{rred}{\textbf{Nick and Warrick investigate the murder of a woman who died in a car crash. The woman's husband, Greg Colletti, is the suspect.}}
  \\\hline
  \multicolumn{1}{c}{\longtv} \\\hline
  Grissom and Catherine
  investigate when a man is found dead in a dumpster.  \textcolor{rred}{\textbf{They soon
  discover a lot more went on in the kitchen than cooking.}}  Meanwhile
  Nick and Sara are called to the scene of a double homicide.  The
  victims are a husband and his wife \textcolor{rred}{\textbf{who were both in the process of
  selling off their rare records.}}  \textcolor{rred}{\textbf{Suspicion quickly falls on the
  wife's ex-boyfriend}}, but the evidence increasingly points to the
  husband.\\\hline
  \multicolumn{1}{c}{\etoe} \\\hline
Grissom, Nick and Catherine investigate when a man is found dead in a dumpster.
Meanwhile a couple are found dead in their home.
Sara and Nick investigate \textcolor{rred}{\textbf{a case involving a record collection.}} \\ \hline
  \multicolumn{1}{c}{\aplan} \\\hline
Grissom and Catherine investigate when a man is found dead in the
garbage.  Their investigation leads to \textcolor{rred}{\textbf{Jackpot's pretzels as the
killer.}}  Meanwhile Sara and Nick are called to a double murder when a
husband and his wife are found dead in their home.  When the husband
and wife were found dead, their extensive record collection was also
missing.\\\hline
\multicolumn{1}{c}{\iterative} \\\hline
Grissom, Catherine and David investigate when a man is found dead in a dumpster.
The man had been eating at a restaurant called Aunt Jackpot's Pretzels.
They discover that he ate himself to death.
Meanwhile Nick and Sara look into the disappearance of a husband and wife who are found dead in their home.
Also missing is a record collection that the husband had been collecting.
They discover that the wife's neck was slashed in the attack.
CSIs track down Missy Halter, a woman who helped them find the
records.\\\hline
\end{tabular}}
  \vspace{-0.1cm}
\caption{\label{tab:system:output} System output and reference summary
  for SummScreen-FD (CSI~S6.E9, ``Dog Eat Dog'').  Propositions which
  are not grounded to the input are in
  \textcolor{rred}{\textbf{red}}. Generated questions from blueprint models are
  not shown due to space constraints.}
    \vspace{-0.5cm}
\end{table}

\paragraph{Blueprint Models and Grounding}

The \etoe\ and \iterative\ variants are significantly better than
\aplan\ in generating texts grounded to their predicted blueprints
(see ground. column in Table~\ref{table:merged-results}).
This is because both models generate text \emph{conditioned} on their
blueprints; \etoe\ first predicts blueprint $b$ and then continues
to generate output $s$ using both $b$ and the input, whereas,
\iterative\ plans and generates one sentence at a time as
$b_i;s_i$. This is not the case with \aplan\ which generates~$s$
conditioned on answer spans only. 
 \etoe\ performs slightly better than \iterative\ on AQuaMuSe 
and WikiCatSum  (differences are not statistically significant)
but struggles on SummScreen-FD, where summaries are longer with more
facts/propositions, requiring inference over long-range dependencies,
and common sense reasoning.  \iterative\ seems the best option for
grounded generation without sacrificing informativeness
(\iterative\ is most informative amongst blueprint models on
SummScreen-FD, second best on AQuaMuSe, and third best on WikiCatSum).

\paragraph{\iterative\ Is Most Faithful Model} 

As far as faithfulness is concerned, \iterative\ performs consistently
better than \etoe\ and \aplan, as well as \tv\ and \longtv\ models
where text is generated from scratch without any planning (pairwise
differences between \iterative\ and comparison systems are all
significant with the exception of \etoe\ on AQuaMuse). On
SummScreen-FD, \iterative\ brings large gains on faithfulness without
sacrificing informativeness (both in terms of Rouge and QA-F1). The
ANLI score for \iterative\ is~20.84, whereas it is below~10 for
\etoe\ and \aplan. \etoe\ outperforms \longtv\ on AQuaMuSe and
WikiCatSum, but gains are smaller compared to \iterative.

We show examples of system output in Table~\ref{tab:system:output},
highlighting propositions which are \emph{not} grounded to the input
in \textcolor{rred}{red}. \etoe\ summaries are shorter which is
somewhat expected; the model has to decode both the plan and the
summary and in cases where the blueprint is large (e.g., in
\mbox{SummScreen-FD}), there is no more room to decode the
summary. \aplan\ is more verbose, however, the plan (a sequence of
answer spans) is less detailed and as a result the summary less
accurate (Jackpot's pretzels is a restaurant, not a
killer). \iterative\ contains many details in the summary, more than
the reference, which are \emph{not} hallucinations.
Both \textsc{r2t-Bart} and \longtv\ are rather loose with the facts and generate multiple hallucinations.

\paragraph{Blueprint Models are Controllable}

Our conceptualization of text plans as QA pairs brings inherent
controllability to the generation process. By changing the blueprint,
we can control content selection (i.e., what to say) and planning
(i.e., in what order) without retraining the model or introducing
additional control mechanisms. 
We provide an example in Table~\ref{tab:blueprint:edit} where the plan
predicted by the \etoe\ model has been edited to render it more
coherent and factual. As can be seen, the model is able to change its
output according to the modified plan. Another example is shown in
Table~\ref{tab:blueprint:short} where the output is rendered shorter
by removing QA pairs from the predicted plan. 

\newcommand\crule[3][black]{\textcolor{#1}{\rule{#2}{#3}}}

\begin{table}[t]
\begin{scriptsize}
\begin{tabular}{|lp{6.5cm}|} \hline
\rowcolor{DarkBP}  & \\ 
  \rowcolor{DarkBP} Q$_{1}$: & What breed existed but is \sout{no longer extinct?}
\raisebox{.3cm}[0pt]{\hspace*{-2cm}\textbf{\textcolor{rred}{now extinct?}}}\\
\rowcolor{DarkBP} A$_{1}$: & Old English Bulldogs \\
\rowcolor{DarkBP} \sout{Q$_{2}$:} & \sout{Along with the Old English Bulldog and Toy Bulldog, what
  breed was considered extinct at the end of the 19th century?}
\textbf{\textcolor{rred}{What was the Old English Bulldog bred for?}}\\
\rowcolor{DarkBP} \sout{A$_{2}$:}& \sout{Bullenbeisser}
\textbf{\textcolor{rred}{Fighting in public arenas}} \\ \hline \hline
\multicolumn{2}{|@{}l@{}|}{~Old English Bulldogs refers to a breed of dog
  that once existed but}\\
\multicolumn{2}{|@{}l@{}|}{~is no extinct. At the end of the 19th century, three breeds
  -- the Old}\\
\multicolumn{2}{|@{}l@{}|}{~English Bulldog, Toy Bulldog, and
  Bullenbeisser -- were considered} \\
\multicolumn{2}{|@{}l@{}|}{~extinct.} \\\hline \hline
\multicolumn{2}{|@{}l@{}|}{\textbf{~\textcolor{rred}{Old English Bulldogs refers to a breed of dog
  that existed but is}}} \\
\multicolumn{2}{|@{}l@{}|}{\textbf{~\textcolor{rred}{now extinct. It was bred for fighting in public arenas.}}} \\ \hline
\end{tabular}
\end{scriptsize}
\caption{\label{tab:blueprint:edit} Example of plan/summary generated
  by our \etoe\ blueprint model as  answer to the question ``What is the
  difference between an Old English Bulldog and an English Bulldog?''
  (AQuaMuse test set); user edits to the plan and updated summary are shown in
  \textbf{\textcolor{rred}{red}}. }
\end{table}

\newcommand{\legendbox}[1]{%
  \textcolor{#1}{\rule{\fontcharht\font`X}{\fontcharht\font`X}}%
}

\begin{table}[t]
  \centering 
\begin{scriptsize}
  \begin{tabular}{|lp{6.5cm}@{\hspace*{-.4ex}}|} \hline
\rowcolor{DarkBP} Q$_{1}$: &  Hinduism is an Indian religion and what else? \hfill A$_{1}$:~dharma \\
\rowcolor{DarkBP} Q$_{2}$: &  Hinduism is a way of what? \hfill A$_{2}$:~life \\
\rowcolor{rred} Q$_{3}$: &  Hinduism has been called what in the world? \hfill A$_{3}$:~the~oldest religion \\
 \rowcolor{rred} Q$_{4}$: &  Who call Hinduism Sanatana Dharma? \hfill A$_{4}$:~some practitioners\\
\rowcolor{rred} Q$_{5}$: &  What does Sanatana Dharma mean? \hfill A$_{5}$:~the eternal tradition \\
\rowcolor{rred} Q$_{6}$: &  Scholars regard Hinduism as a fusion of various Indian cultures
and what else? \hfill A$_{6}$:~traditions  \\
\rowcolor{rred} Q$_{7}$: &  Hinduism has no founder and what else? \hfill A$_{7}$:~diverse roots\\
\rowcolor{rred} Q$_{8}$: &  What started to develop between 500 BCE and 300 CE? \hfill A$_{8}$:~This Hindu synthesis\\
\rowcolor{rred} Q$_{9}$: &  When did the Vedic period end? \hfill A$_{9}$:~500 BCE \\\hline\hline
\multicolumn{2}{|@{\hspace*{0.5ex}}l@{\hspace*{0.6ex}}|}{Hinduism is an Indian religion and dharma, or a way of life,
  widely} \\
\multicolumn{2}{|@{\hspace*{0.5ex}}l@{\hspace*{-.1ex}}|}{practiced in the Indian subcontinent. Hinduism has been
  called the}\\ 
\multicolumn{2}{|@{\hspace*{0.5ex}}l@{\hspace*{-.1ex}}|}{
    oldest religion in the world, and some practitioners and scholars re-}\\
\multicolumn{2}{|@{\hspace*{0.5ex}}l@{\hspace*{-.1ex}}|}{fer to it
     as Sant\={a}na Dharma, ``the eternal
     tradition'', or the ``eternal}\\
 \multicolumn{2}{|@{\hspace*{0.5ex}}l@{\hspace*{-.1ex}}|}{way'', beyond human
   history. Scholars regard Hinduism as a fusion} \\
 \multicolumn{2}{|@{\hspace*{0.5ex}}l@{\hspace*{-.1ex}}|}{or synthesis of various Indian
   cultures and traditions, with diverse} \\
\multicolumn{2}{|@{\hspace*{0.5ex}}l@{\hspace*{-.1ex}}|}{roots and no founder. This
  "Hindu synthesis" started to develop} \\
\multicolumn{2}{|@{\hspace*{0.5ex}}l@{\hspace*{-.1ex}}|}{between 500 BCE and 300 CE,
  following the Vedic period (1500}\\
\multicolumn{2}{|@{\hspace*{0.5ex}}l@{\hspace*{-.1ex}}|}{BCE to 500 BCE).}  \\ \hline \hline
\multicolumn{2}{|@{\hspace*{0.5ex}}l@{\hspace*{-.1ex}}|}{\textbf{\textcolor{rred}{Hinduism is an Indian religion and
    dharma, or a way of life,}}} \\
 \multicolumn{2}{|@{\hspace*{0.5ex}}l@{\hspace*{-.1ex}}|}{\textbf{\textcolor{rred}{widely practiced in the Indian
     subcontinent.}}} \\\hline
\end{tabular}
\end{scriptsize}
\caption{\label{tab:blueprint:short} Example of plan/summary generated
  by the \etoe\ blueprint model as answer to the question ``What
  section of the world or country is hinduism usually found in?
  (AQuaMuse test set); the part of the plan which is removed by the
  user is highlighted in \legendbox{rred}\legendbox{red}; the shorter
  summary generated from the elided plan is shown in
  \textbf{\textcolor{rred}{red}}.}
\end{table}

\begin{table}[t!]
\center
 \footnotesize
 {\setlength{\tabcolsep}{1.5pt}%
\begin{tabular}{ll | c c | c | c } \hline
&& \multicolumn{2}{c|}{Rouge (RLSum)} &  QA-F1 & ANLI  \\ 
& \multicolumn{1}{c|}{Models} & \multicolumn{1}{c}{summ.} & \multicolumn{1}{c|}{bluepr.} & \multicolumn{1}{c|}{inform.} & \multicolumn{1}{c}{entail.} \\ \hline 
&\cellcolor{DarkBP}\textcolor{black}{\etoe} & \hspace*{1ex}{58.96}$^{\dagger}$ & 54.16$^{\dagger}$ & 34.59 & 83.64$^{\dagger}$  \\
&\hspace{0.2cm} \cellcolor{MidBP}\textcolor{black}{+drop} & \hspace*{1ex}58.74$^{\dagger}$ & 52.71$^{\dagger}$ & 34.64 & 83.98$^{\dagger}$ \\
&\cellcolor{DarkBP}\textcolor{black}{\aplan} & \hspace*{1ex}59.24$^{\dagger}$ & \hspace*{-1ex}\textbf{54.88} & \hspace*{1ex}\textbf{37.87}$^{\dagger}$ & 83.37$^{\dagger}$ \\
&\hspace{0.2cm} \cellcolor{MidBP}\textcolor{black}{+drop} & \textbf{59.25} & 53.79$^{\dagger}$ & 37.57 & 84.98$^{\dagger}$ \\
&\cellcolor{DarkBP}\textcolor{black}{\iterative} & 56.48 &  52.92$^{\dagger}$ & 36.04 & 84.77$^{\dagger}$ \\
& \hspace{0.2cm} \cellcolor{MidBP}\textcolor{black}{+drop} & 56.30 & \hspace*{-1ex}46.62 & 36.03 & {85.50}$^{\dagger}$ \\
\raisebox{4ex}[0pt]{\begin{sideways}AQuaMuse\end{sideways}} &\hspace{0.2cm} \cellcolor{LightBP}\textcolor{black}{+Q1} & 55.57 & 52.29$^{\dagger}$ & 34.97 & \hspace*{-1ex}\textbf{85.74}\\ 
\hline 
&\cellcolor{DarkBP}\textcolor{black}{\etoe} & 39.78 & 45.40 &  \hspace*{1ex}19.22$^{\dagger}$ & \hspace*{-1ex}60.56\\
&\hspace{0.2cm} \cellcolor{MidBP}\textcolor{black}{+drop} & 37.35 & 36.00 & 17.74 & \hspace*{-1ex}64.38  \\
&\cellcolor{DarkBP}\textcolor{black}{\aplan} & \textbf{42.34}  & \textbf{46.95} &\textbf{20.55} & \hspace*{-1ex}52.09 \\
&\hspace{0.2cm} \cellcolor{MidBP}\textcolor{black}{+drop}  & 40.65 & 36.45 & \hspace*{1ex}19.29$^{\dagger}$ & \hspace*{-1ex}56.71 \\
&\cellcolor{DarkBP}\textcolor{black}{\iterative} & 39.53 & \hspace*{1ex}46.35$^{\dagger}$ & 18.75 &\hspace*{-1ex}64.37 \\
&\hspace{0.2cm} \cellcolor{MidBP}\textcolor{black}{+drop}  & 38.80 & 41.78  & 18.14 & \hspace*{-1ex}\textbf{66.56} \\
\raisebox{3ex}[0pt]{\begin{sideways}WikiCatSum\end{sideways}}&\hspace{0.2cm} \cellcolor{LightBP}\textcolor{black}{+Q1} & 38.80 & 43.81 & 18.91 & 65.42$^{\dagger}$ \\
\hline 
&\cellcolor{DarkBP}\textcolor{black}{\etoe} & 28.22 & 34.49  & 6.13 & 8.53\\
&\hspace{0.2cm} \cellcolor{MidBP}\textcolor{black}{+drop}  & 25.50 & 23.22 & 5.28 & \hspace*{-1ex}10.00 \\
&\cellcolor{DarkBP}\textcolor{black}{\aplan} & \textbf{31.88} & 35.23 & \hspace*{1ex}7.06$^{\dagger}$ & 9.57  \\
&\hspace{0.2cm} \cellcolor{MidBP}\textcolor{black}{+drop}  & \hspace*{1ex}30.40$^{\dagger}$ & 24.75 & 6.10 & \hspace*{-1ex}12.23\\
&\cellcolor{DarkBP}\textcolor{black}{\iterative} & \hspace*{1ex}29.86$^{\dagger}$ & \textbf{39.36}  & \textbf{7.55} & 20.84$^{\dagger}$\\
& \hspace{0.2cm} \cellcolor{MidBP}\textcolor{black}{+drop} &  28.10 & 32.25 & 6.43 & {24.12}$^{\dagger}$  \\
\raisebox{2ex}[0pt]{\begin{sideways}SummScreen\end{sideways}}&\hspace{0.2cm} \cellcolor{LightBP}\textcolor{black}{+Q1} & 27.94 & 29.47 & 6.63 & \hspace*{-1ex}\textbf{24.80} \\
\hline
\end{tabular}
}
\caption{Controllability results on the AQuaMuSe, WikiCatSum and
  SummScreen-FD test sets. Lighter blue color means more control. Best
  results for each metric are
  \textbf{boldfaced}. Scores that are \uline{not} significantly different
  (using paired bootstrap resampling; \mbox{$p<0.05$}) from the best
  score for each column are marked with a dagger (${\dagger}$). \label{table:controlled-results} }
 \end{table}

We are also able to control the faithfulness of predicted summaries as
follows. We take the predicted plan and remove question-answer pairs (\etoe,
\iterative) or answer spans (\aplan) that cannot be answered based on
the input. We then prompt our decoder with the modified plan and
generate a new summary (or sentence for \iterative).
In Table~\ref{table:controlled-results}, we quantitatively evaluate
+drop variants which are controlled for faithfulness against vanilla
blueprint models. We observe improvements in entailment scores across
the board (see column entail. in the table), with the \iterative+drop
performing best. Improvements on abstractive datasets (WikiCatSum and
SummScreen-FD) are larger compared to AQuaMuSe which is mostly
extractive (see Table~\ref{tab:dataset}). The minor drop in Rouge and
informativeness is somewhat expected as the models now zoom in on
information they can \emph{reliably} talk about, improving the
consistency of the output.

\begin{table}[t]
  \setlength\extrarowheight{-5pt}
\resizebox{\linewidth}{!}{
{\setlength{\tabcolsep}{0pt}%
  \begin{tabular}{@{}p{8.2cm}@{~~}p{5cm}} \hline
    \multicolumn{2}{c}{\iterative} \\ \hline
\rowcolor{DarkBP} \textcolor{black}{Q$_{1}$: In what country was Abraham
  Verghese born?} & \cellcolor{white}Abraham Verghese was born in\\
\rowcolor{DarkBP}  \textcolor{black}{A$_{1}$: Ethiopia}   &
\cellcolor{white}Ethiopia, while his parents, \\
\rowcolor{DarkBP} \textcolor{black}{Q${_2}$: From what country did
  Verghese's parents come?} &\cellcolor{white} teachers from Kerala, India. \\
\rowcolor{DarkBP} \textcolor{black}{A$_{2}$: India}  &
\cellcolor{white}  lived there. \\& \cellcolor{white} \\
\rowcolor{DarkBP} \textcolor{black}{Q$_{3}$: Cutting for Stone is set in
  Ethiopia and what other country?} & \cellcolor{white}The main
characters in this gripping family saga, set in Ethiopia     \\
   \rowcolor{DarkBP} \textcolor{black}{A$_{3}$: the United States}
   &  \cellcolor{white} and the United States, are two  \\
\rowcolor{DarkBP} \textcolor{black}{Q$_{4}$: What are the main characters
  in Cutting for Stone?} & \cellcolor{white} physicians from India.  \\
   \rowcolor{DarkBP} \textcolor{black}{A$_{4}$: two physicians} &
    \cellcolor{white} \\ \\
\rowcolor{DarkBP} \textcolor{black}{Q$_{5}$: In what year did Cutting for
  Stone win the Man Booker Prize?} & \cellcolor{white} Cutting for
Stone won the Man Booker Prize in 2007.\\
\rowcolor{DarkBP} \textcolor{black}{A$_{5}$: 2007}  &
\cellcolor{white} \\ \\
\rowcolor{DarkBP} \textcolor{black}{Q$_6$: Who is the author of
  Cutting for Stone?} & \cellcolor{white}The author is Ethiopian-born \\
\rowcolor{DarkBP} \textcolor{black}{A$_6$: Ethiopian-born Verghese}
&  \cellcolor{white}{ Verghese.}
\\\hline

\multicolumn{1}{c}{} \\ \hline
\multicolumn{2}{c}{\iterative+Q1} \\  \hline
\rowcolor{DarkBP} \textcolor{black}{Q$_{1}$: In what country was Abraham Verghese born?} & \cellcolor{white}Abraham Verghese was born in \\
 \rowcolor{DarkBP} \textcolor{black}{A$_{1}$: Ethiopia} &
 \cellcolor{white} Ethiopia.\\\\
\rowcolor{DarkBP} \textcolor{black}{Q$_{2}$: From what country did
  Verghese's parents come?} & \cellcolor{white} His parents were
teachers \\
\rowcolor{DarkBP} \textcolor{black}{A$_{2}$: India} &
\cellcolor{white} from Kerala, India.\\ \\
\rowcolor{DarkBP} \textcolor{black}{Q$_{3}$: Cutting for Stone is
  set in Ethiopia and what other country?} & \cellcolor{white} Cutting
for Stone is a gripping family saga, set in Ethiopia.\\
\rowcolor{DarkBP} \textcolor{black}{A$_{3}$: the United States} &
\cellcolor{white}  \\ \cellcolor{white}\\
\rowcolor{DarkBP} \textcolor{black}{Q$_{4}$: What are the main
  characters in Cutting for Stone?} & \cellcolor{white}The main
characters are two\\
 \rowcolor{DarkBP} \textcolor{black}{Q$_{4}$: two physicians} &
 \cellcolor{white}  physicians  from India. \\\hline
\end{tabular}}}
\caption{\label{tab:system:output:controlled} System output from \iterative\ and \iterative+Q1 generating WikiCatSum abstract on ``Abraham Verghese.''
}
\end{table}

Finally, we also experiment with creating simple summaries, by forcing
the \iterative\ model to generate from a \emph{single} question-answer
pair on each iteration (see~+Q1 variant in
Table~\ref{table:controlled-results}). In the example shown in
Table~\ref{tab:system:output:controlled}, \iterative+Q1 produces
simple summary sentences, each focusing on a single information
element. Interestingly, as far as the \iterative\ model is concerned,
+Q1 variants are as faithful as +drop ones even if they do not
explicitly control for faithfulness (across datasets the differences
between the two models are not statistically significant). This
suggests that controlling for simplicity might be sufficiently to
reduce hallucinations, however, at the expense of informativeness
(Rouge scores for +Q1 variants tend to be significantly worse compared
to +drop counterparts).

Most of the controllability cases we illustrate here are fully
automatic and could be conceptualized as system flags which users
select according to requirements (e.g., low tolerance for
hallucinations, shorter summaries for small screen displays).  Another
potential use case would be to generate summaries for a set of
questions provided by the user. Their input might be articles
retrieved as an answer to a query, or in an educational context
several chapters on a topic (e.g.,~cell biology). However, we leave
this to future work.

\begin{table}[t!]
\center
\footnotesize
\begin{tabular}{@{}l @{~~}|@{~} c@{~} c@{~} c@{}} \hline
&  \multicolumn{3}{c}{Rouge (RLSum)} \\
\multicolumn{1}{c|}{\etoe}  & summary & blueprint & both \\ \hline
QA Plan, Rheme, Covg, Sorted & 48.75 & 39.06 & 44.31 \\
AQ Plan, Rheme, Covg, Sorted & \textbf{50.86} & \textbf{39.95}  & \textbf{45.60} \\
\hspace{0.4cm} $-$Sorted, Random & 50.79 & 36.08 & 43.43 \\
\hspace{0.4cm} $-$Rheme & 47.16 & 40.70 & 44.19 \\
\hspace{0.4cm} $-$Coverage & 47.02 & 41.37 & 44.79 \\
\hspace{0.4cm} $-$Rheme, $-$Coverage  & 18.05 & \textbf{42.54} & 40.90\\
\hline
\end{tabular}
\caption{\etoe\ model trained on AQuaMuSe with different selection and
  sorting  (validation set). \label{table:selection-sorting-ablations}}
\end{table}

 \subsection{Ablation Studies}
 \label{sec:ablations}

 As described in Section~\ref{subsec:annotation}, we construct
 blueprint annotations using the Rheme- and Coverage-based selection
 strategies. Table~\ref{table:selection-sorting-ablations} presents
 various ablations which provide rationales for these annotation
 choices. For the sake of brevity, we report experiments with the
 \etoe~model trained (for 50,000 steps) on AQuaMuSe. We observe very
 similar trends on the other two datasets. As can be seen, it is
 empirically better to form blueprints from answer-question pairs
 rather than predicting the questions first and then their answers
 which is more natural (at least to humans).
We further assessed whether sorting the QA pairs based on how they
appear in the summary matters by defaulting to a random ordering (see
$-$Sorted in the table).
Removing either Rheme or Coverage has a small negative impact on the
summaries but not their blueprints, while removing them both is
detrimental to summary quality, while the absence of Sorting mostly
affects the quality of the blueprint.  It is not surprising that
sorting is most important to generating a blueprint with correctly
ordered propositions.

 \begin{table}[t!]
\center
\small
\begin{tabular}{l|@{}c@{}|@{}c@{}|@{}c@{}} \hline
 \multicolumn{1}{c}{}& \multicolumn{3}{c}{summary quality $\uparrow$}  \\
\multicolumn{1}{@{}c|}{Models}     &  \multicolumn{1}{c@{}|@{}}{AQuaMuse} &
 \multicolumn{1}{@{}c@{}|@{}}{WikiCatSum} &\multicolumn{1}{@{}c@{}}{SummScreen} \\ \hline
\textsc{SiBERT} & \hspace*{-1ex}0.12$^{}$ & --- & --- \\
\textsc{r2t-Bart} & --- & --- & 0.10$^{}$ \\
\longtv & \hspace*{-.7ex}\textbf{0.49}$^{}$ & 0.35$^{\dagger}$& \hspace{1ex}0.42$^{\dagger}$ \\
\cellcolor{DarkBP}{\etoe} & 0.39$^{\dagger}$& \hspace*{-.7ex}\textbf{0.40}& 0.26$^{}$ \\
\cellcolor{DarkBP}{\aplan} &0.39$^{\dagger}$& 0.37$^{\dagger}$& \hspace*{1ex}0.39$^{\dagger}$ \\
\cellcolor{DarkBP}{\iterative} &\hspace*{-.7ex}0.30$^{}$&0.28$^{\dagger}$& \hspace*{.2ex}\textbf{0.44}$^{}$ \\
\hspace{0.2cm} \cellcolor{MidBP}+drop
&\hspace*{-.7ex}0.31$^{}$&0.27$^{\dagger}$& \hspace*{1ex}0.39$^{\dagger}$ \\
\hline
\end{tabular}
\caption{Proportion of times each system was ranked best for summary
  quality (on AQuaMuse, WikiCatSum, and SummScreen test sets). Best
  results for each task are \textbf{boldfaced}.  Systems in each
  column are marked with~$\dagger$ when they are \uline{not}
  significantly different from the best system; unmarked pairwise
  differences frin the best system are significant ($p<0.01$; using
  Friedman's ANOVA test (with post-hoc Wilcoxon signed-rank test,
  Bonferroni corrected for multiple
  comparisons). \label{table:heval-results} }
\end{table}

\section{Human-based Evaluation}

In addition to automatic evaluation, we conducted three human-based
studies which assessed different dimensions of output quality. Wishing
to avoid well-documented
issues\footnote{\url{https://stanforddaily.com/2020/06/21/}}
with automated bots on Amazon Mechanical Turk and crowdworkers running
through HITs as quickly as possible without paying attention to the
tasks, we used a few trained annotators. They were given task-specific
instructions and went through several pilots to iron out disagreements
on edge cases.\footnote{We will release our instructions and
  annotation templates together with our data and models.}

\subsection{Summary Quality}
Our first study assessed overall {\em summary quality}.
Specifically, we asked our annotators to select the best
among three system summaries taking into account how much they
deviated from the reference in terms of \emph{informativeness} (are
the summaries on topic or emphasize irrelevant details?) and overall
\emph{fluency}. We adapted the definition of fluency provided in
\citet{howcroft-etal-2020-twenty}:  does the text ‘flow well’ or is it
a  sequence of unconnected parts? 

We conducted our annotation study on 100~instances, each randomly
sampled from AQuaMuse, WikiCatSum, and SumScreen. We collected ratings
from three annotators (after two rounds of pilot studies to improve
agreement) for the output of seven systems. Overall, we obtained 100
(instances) x 3 (datasets) x 6 (systems) x 3 (annotators) = 5,400
annotations.  Annotator agreement was~97.11\%. Our results are
presented in Table~\ref{table:heval-results}.  We report on percentage
of times each system was ranked best.

In general, we observe that \longtv\ and blueprint models based on it
are perceived as significantly better than previous state-of-the-art
models (i.e., \textsc{SiBERT} and \textsc{r2t-Bart}). On AQuaMuse,
\longtv\ is rated overall best, followed by \etoe\ and \aplan
(however, differences between them are not statistically
significant). On WikiCatSum, \etoe\ is rated best bus is not
significantly different compared to the other models. On SummScreen,
our \iterative\ variant is rated best followed by \longtv. These
results mirror the difficulty of the task (see 
Table~\ref{tab:dataset}), the longer the input/output, the better
\iterative\ performs.

\subsection{Blueprint Quality}


We further evaluated the predicted plans more directly. Participants
were shown QA blueprints and asked to assess whether they tell a
coherent story (are they all relevant and ordered comprehensively?)
using a 3-point scale (where 3 is best and 1 is worst). They were also
asked to evaluate whether the plans have redundant QA pairs; a QA pair
is redundant if it does not add new information to the plan. We
collected judgments for the same instances used in our summary quality
evaluation from three annotators whose overall agreement was~97.87\%.
Obtained a total of 100 (instances) x 3 (datasets) x 5 (systems) x 3
(raters) = 4,500 annotations.

\begin{table}[t!]
\small
  \begin{tabular}{l@{~}|@{~}c@{}c@{}|@{}c@{}c@{}|@{}c@{}c@{}} \hline
\multicolumn{1}{@{\hspace*{0cm}}c}{Blueprint}   & \multicolumn{2}{c}{AQuaMuse} &
  \multicolumn{2}{@{~}c@{}}{WikiCatSum}& \multicolumn{2}{@{~}c@{}}{SummScreen} \\
\multicolumn{1}{@{~}c|}{Models}     &  Coh$\uparrow$ & Red$\downarrow$ & Coh$\uparrow$ & Red$\downarrow$ & ~Coh$\uparrow$ & Red$\downarrow$ \\\hline
\cellcolor{DarkBP}\textcolor{black}{\etoe}
& \textbf{2.83}$^{}$ & \hspace{.2ex}\textbf{8.30}$^{}$
& \textbf{2.94}$^{}$ & \textbf{4.70}$^{}$&  \hspace*{-1ex}\textbf{2.43}$^{}$ & 49.3$^{\dagger}$ \\
\cellcolor{DarkBP}\textcolor{black}{\aplan} & 2.58$^{}$
& 16.00$^{\dagger}$ & \hspace{.7ex}2.84$^{\dagger}$ & \hspace{.7ex}6.30$^{\dagger}$ &  2.28$^{\dagger}$ &
\hspace*{-1ex}\textbf{33.0}$^{}$ \\
\cellcolor{DarkBP}\textcolor{black}{\iterative} & 2.51 &
40.00$^{}$& 2.64$^{}$ & \hspace*{-.7ex}31.00$^{}$ &\hspace*{-1ex}2.09$^{}$ & \hspace{-1ex}81.7$^{}$  \\
\hspace{0.2cm} \cellcolor{MidBP}{+drop}  &
2.56 & 36.70$^{}$ & 2.66$^{}$ &  \hspace{-.7ex}27.30$^{}$ &
\hspace*{-1ex}2.28  & \hspace*{-1ex}63.7 \\ \hline
Gold  & \hspace{.7ex}2.85$^{\dagger}$ &  \hspace{2ex}7.00$^{\dagger}$ & 2.67$^{}$  &  \hspace{-.2ex}14.30$^{\dagger}$  &  \hspace*{-.5ex}2.02$^{}$ & \hspace{-1ex}95.7  \\ \hline
\end{tabular}
\caption{Blueprint quality human evaluation on AQuaMuse, WikiCatSum,
  and SummScreen-FD test sets. Mean scores for coherence (Coh; higher
  is better) and proportion of QA pairs deemed redundant (Red; lower
  is better).  Best results for each task are
  \textbf{boldfaced}. Systems in each column are marked with~$\dagger$
  when they are \uline{not} statistically significant from the best
  system; unmarked pairwise differences from the best system are
  significant ($p<0.01$; using a Friedman's ANOVA test with post-hoc
  Wilcoxon signed-rant test, Bonferroni corrected for multiple
  comparisons).
  \label{table:coh:eval-results} }
\end{table}

Table~\ref{table:coh:eval-results} shows the results of this study. We
report mean scores per dataset for all blueprint models. As an upper
bound, we further elicited annotations for blueprints automatically
created from \emph{gold} standard reference summaries (see row Gold in
the table).  \etoe\ generates the most coherent blueprints:
differences between \etoe\ and all comparison systems are
statistically significant with the exception of the gold standard. This
is not surprising, since all QA pairs in \etoe\ are generated
together, whereas in \aplan\ the spans and their corresponding
questions are generated separately. \iterative\ only generates QA
pairs for a sentence at a time and thus we would not expect it to be
more coherent than models which generate a global document plan.  With
regard to redundancy, \iterative\ blueprints are generally most
redundant, which is again down to not having a global view of
previously generated QA pairs.  \iterative\ further underscores issues
with our question generation technology which is far from perfect, for
example, several QA pairs are different on the surface but actually
semantically equivalent, however, we have no means of detecting this
without robust coreference resolution.


\subsection{Blueprint Grounded Generation}

We next examine whether model summaries are grounded to their
blueprints. Specifically, we asked our annotators to decide whether
each QA pair in the blueprint is mentioned in the summary, and report
the number of times it isn't. Ideally, we would like the summary to
follow the blueprint as closely as possible. For QA pairs mentioned in
the summary, we further asked our annotators to highlight whether the
intent of the question was preserved or contradicted (we report the
number of contradictions). Finally, we also asked participants to
decide whether the summary has \emph{additional} information which
cannot be found in its blueprint, using a \mbox{3-point} scale (where
3 is for summaries with lots of new information and 1 is for summaries
with no new information). We elicited annotations for blueprint
models, and, as an upper bound, for gold summaries and blueprints
extrapolated from them. We obtained  100 (instances) x 3
(datasets) x 5 (systems) x 3 (raters) = 4,500 judgments. 

\begin{table*}[t!]
\center
\begin{tabular}{l|c@{~~}c@{~}c@{~}|@{~}c@{~}c@{~}c@{~}|@{~}c@{~}c@{~}c}
 & \multicolumn{3}{c}{AQuaMuse} & \multicolumn{3}{c}{WikiCatSum} &
\multicolumn{3}{c}{SummScreen} \\ 
Models  & Absent$\downarrow$ & Contra$\downarrow$ & NewInfo$\downarrow$ &  Absent$\downarrow$ & Contra$\downarrow$ & NewInfo$\downarrow$ & Absent$\downarrow$ &
  Contra$\downarrow$ & NewInfo$\downarrow$  \\\hline
\cellcolor{DarkBP}\textcolor{black}{\etoe} & 6.30$^{\dagger}$  & 4.10$^\dagger$  & 1.79$^{\dagger}$ &
2.00$^\dagger$  & \hspace{-.7ex}\textbf{0.40}  & \hspace{-1ex}\textbf{1.74}  & 26.60 & \hspace{1ex}5.30$^\dagger$ & \hspace{.8ex}1.76$^\dagger$ \\
\cellcolor{DarkBP}\textcolor{black}{\aplan} &  \hspace{-1ex}13.20$^{\dagger}$ & \hspace*{-2ex}13.60 & 2.06
& 5.80$^\dagger$  & 3.70$^\dagger$  & 1.91$^\dagger$  & 63.00 & 8.70  & 2.60 \\
\cellcolor{DarkBP}\textcolor{black}{\iterative} & 5.00$^{\dagger}$  & 2.50$^\dagger$  &
\hspace{-1ex}\textbf{1.77}  & \hspace{-1ex}\textbf{2.80}  & 2.90$^\dagger$  & {1.90}$^{\dagger}$  &  \textbf{12.00} & \textbf{2.80} & \textbf{1.53} \\
\hspace{0.2cm} \cellcolor{MidBP}\textcolor{black}{+drop} &
\hspace{-1ex}\textbf{4.90}  & \textbf{2.30}$^\dagger$  &
1.86$^\dagger$  & 3.60$^{\dagger}$
& 2.10$^{\dagger}$  & \hspace{-1ex}\textbf{1.74} & \hspace*{1ex}14.20$^\dagger$ & \hspace{1ex}3.60$^{\dagger}$ & 1.89 \\
\hline
 Gold & \hspace*{-.6ex}3.60$^{\dagger}$ & 2.40$^{\dagger}$
 & 1.88$^{\dagger}$  & 1.60$^{\dagger}$ &  \hspace*{-1ex}1.80 &  \hspace*{-.8ex}2.02  & \hspace{1.5ex}4.50 & 3.90 & 1.39  \\ \hline
\end{tabular}
\caption{Human evaluation results for blueprint grounded generation on
  AQuaMuse, WikiCatSum, and SummScreen-FD test sets. Proportion of QA
  pairs \emph{not} mentioned in the summary (Absent; lower is better);
  proportion of QA pairs with information contradictory to the summary
  (Contra; lower is better), and mean scores for new information
  present in the summary (NewInfo; lower is better).  The best results
  for each task are \textbf{boldfaced}.  Systems in each column are
  marked with~$\dagger$ when they are \uline{not} statistically
  significant from the best system; unmarked pariwise differences from
  the best system are significant ($p<0.01$; using a Friedman's ANOVA
  test with post-hoc Wilcoxon signed-rant test, Bonferroni corrected
  for multiple comparisons).  \label{table:heval-grounding} }
\end{table*}





The results of our grounding experiments are summarized in
Table~\ref{table:heval-grounding}. Across datasets, we observe that
\iterative\ summaries are most grounded. \iterative\ blueprints have
the least number of questions that are absent from or contradict their
generated texts. \iterative\ summaries also display the least amount
of new information in relation to their blueprints.  \iterative+drop
is slightly less grounded compared to \iterative, however, this is not
entirely surprising since we prompt the \iterative\ model with
externally modified blueprints (see \iterative+drop in
Table~\ref{table:heval-grounding}). Note that \iterative+drop
summaries are deemed more faithful than \iterative\ summaries in
automatic evaluation. The entailment scores improve for all three
datasets (see Table~\ref{table:merged-results}).




\section{Conclusion}

In this work we proposed a novel plan-based approach to conditional
generation. 
We conceptualized text plans as a sequence of QA pairs operating as a
proxy for what to say and in what order. We developed
Transformer-based models which generate by conditioning on a global QA
blueprint plan (\etoe, \aplan) or iteratively by planning and
generating one sentence at a time (\iterative).  Experimental results
across three challenging datasets demonstrate that blueprint models
are inherently more informative than vanilla sequence-to-sequence
approaches without a planning component. Amongst the three presented
here (\etoe, \aplan, \iterative), we find that \iterative\ is the best
choice for grounded generation and a promising direction for long-form
generation.

Blueprint models offer several advantages compared to blackbox
generation. Model predictions can be examined, and errors can be
traced back to the blueprint which in turn can reveal whether the
output is informative and faithful to its input.  The formulation of
the blueprint plan as question-answer pairs makes it intuitive and
user-friendly.  We have discussed how blueprint models might be used
in a human-in-the-loop setting, where users interact with and
influence model predictions directly, e.g.,~by editing the blueprint
length and content (as different blueprints lead to different
outputs). In the future, we would like to use blueprints more directly
to advance methods for training language models using reward learning
\cite{Sutton:Barto:2018}, e.g., based on whether the output answers
the blueprint questions. Rather than eliciting expensive human
feedback \cite{NEURIPS2020_1f89885d}, blueprints could provide a
cheaper automatic alternative. Finally, although we focused primarily
on the generation problem in this work, we believe blueprints might
also be useful as a general-purpose approach to retrieving and
organizing important content, especially when faced with many and very
long inputs.

\section*{Acknowledgments}
We thank the action editor, Mark Johnson, and our reviewers, for their valuable feedback. 
The human rating process was managed by Muqthar Mohammad, Kiranmai Chennuru, Ashwin Kakarla and their team, without them this work would not have been possible. Thanks for invaluable support from Sheila de Guia and Suneet Dhingra.

\bibliography{anthology,qud,custom}
\bibliographystyle{style-files/TACL/acl_natbib}









\end{document}